\documentclass[letterpaper]{article} 
\usepackage{aaai25}  
\usepackage{times}  
\usepackage{helvet}  
\usepackage{courier}  
\usepackage[hyphens]{url}  
\usepackage{graphicx} 
\usepackage{natbib}  
\usepackage{caption} 
\usepackage{amsmath}
\usepackage{amsfonts}
\usepackage{amssymb}
\usepackage{algorithm}
\usepackage{algpseudocode}
\usepackage{graphicx}
\usepackage{subcaption} 
\usepackage{amsmath}
\usepackage{amsthm}
\usepackage{xcolor}
\usepackage{newfloat}
\usepackage{listings}
\DeclareCaptionStyle{ruled}{labelfont=normalfont,labelsep=colon,strut=off} 
\floatstyle{ruled}
\newfloat{listing}{tb}{lst}{}
\floatname{listing}{Listing}
\usepackage{soul}

\nocopyright 
\title{Focus Where It Matters: Graph Selective State Focused Attention Networks}

\author{Shikhar Vashistha, Neetesh Kumar}
\affiliations{
    Indian Institute of Technology, Roorkee\\
    shikhar\_v@cs.iitr.ac.in, neetesh@cs.iitr.ac.in
    }

\usepackage{bibentry}
\usepackage{color}

\begin{document}

\maketitle

\begin{abstract}

  Traditional graph neural networks (GNNs) lack scalability and lose individual node characteristics due to over-smoothing, especially in the case of deeper networks. This results in sub-optimal feature representation, affecting the model's performance on tasks involving dynamically changing graphs. To address this issue, we present Graph Selective States Focused Attention Networks (GSANs) based neural network architecture for graph-structured data. The GSAN is enabled by multi-head masked self-attention (MHMSA) and selective state space modeling (S3M) layers to overcome the limitations of GNNs. In GSAN, the MHMSA  allows GSAN to dynamically emphasize crucial node connections, particularly in evolving graph environments. The S3M layer enables the network to adjust dynamically in changing node states and improving predictions of node behavior in varying contexts without needing primary knowledge of the graph structure. Furthermore, the S3M layer enhances the generalization of unseen structures and interprets how node states influence link importance. With this, GSAN effectively outperforms inductive and transductive tasks and overcomes the issues that traditional GNNs experience. To analyze the performance behavior of GSAN, a set of state-of-the-art comparative experiments are conducted on graphs benchmark datasets, including \textit{Cora}, \textit{Citeseer}, \textit{Pubmed} network citation, and \textit{protein-protein-interaction} datasets, as an outcome, GSAN improved the classification accuracy by \textit{1.56\%}, \textit{8.94\%}, \textit{0.37\%}, and \textit{1.54\%} on \textit{F1-score} respectively.
\end{abstract}
\section{Introduction}
    Graph Neural Networks (GNNs) are specifically designed for analyzing and processing graph data as it's difficult to restrict data to \textit{Euclidean} space for all tasks where the relationship, interaction, and state of the nodes carry important information. GNNs capture and process local and global information effectively, allowing them to preserve and exploit relational information in the data. There are several attempts in the literature to utilize graphs to the best capacity by extending conventional neural networks for dealing with the dynamism of graph-structured data \cite{zheng2024surveydynamicgraphneural}. Early works utilized recursive neural networks for dealing with graph data, which were later generalized to GNNs in 2005 and 2009 that can operate on the more generalized graph classes, e.g., cyclic, directed, and undirected \cite{1555942}. GNNs usually propagate the node state until the equilibrium, which was adopted and enhanced using gated recurrent units (GRUs) inside the propagation step by \cite{li2016gated}. Further progress in GNNs advances to spectral and non-spectral approaches.

    The spectral representation of graphs examines graph properties via the adjacency matrix and laplacian matrix. The adjacency matrix \(A\) represents edges connecting nodes. The laplacian matrix, defined as \(L = D - A\), where \(D\) represents the degree matrix which is for understanding the graph's structural connectivity and community dynamics \cite{Shuman2020}. The eigenvalues and eigenvectors of these matrices disclose important characteristics of the graph, such as overall connectivity and the presence of community structures, influencing methods for clustering and dimensionality reduction \cite{Lee2021}. These spectral techniques are required in applications such as social networks and bio-informatics, providing analytical tools for complex graph network analysis \cite{Zhou2022}.

    Non-spectral methods in GNNs are known for spatial or message-passing techniques. These methods include processing nodes and their local neighborhoods without spectral transformation. They use local message passing, where nodes update representations by aggregating features from direct neighbors, capturing graph topology effectively \cite{hamilton2020inductive, Hamilton2020}. These methods include graph convolutional networks (GCNs), which simplify graph processing by focusing on immediate neighbor interactions \cite{Wu2021} and perform first-order approximations of the spectral filters \cite{kipf2017semi}, and graph attention networks (GATs), which use an attention mechanism to dynamically prioritize information from various neighbors \cite{zhang2021inductive}. These methods scale well, handle large and dynamic graphs, and are used in social network analysis and traffic forecasting \cite{Chami2021}. Additionally, graph state space models (GSSMs) integrate graphs into state space models to capture complex variable interactions \cite{zambon2023graphstatespacemodels}. 

        The existing methods have limitations that restrict their applicability in wider domains. For instance, graph-enhanced spatial-temporal aware attention network (GSTAN) excels in spatial-temporal recommendations but lacks versatility across diverse graph-based tasks, which restricts its utility beyond its niche application \cite{cao2023improving}. An \cite{li2024stgmamba} approach integrates selective state space modeling (S3M) to enhance spatial-temporal dynamics but does not utilize attention mechanisms for richer feature extraction across varying graph environments. Similarly, GAT was adopted for task allocation in mobile crowdsensing. Yet, it could be further refined to dynamically adjust node states, enhancing its effectiveness across varying contexts \cite{xu2023intelligent}. A \cite{cao2023improving} method merges geospatial information with graph attention mechanisms effectively; however, it primarily focuses on change detection and might underutilize the potential of dynamic state adjustments that could benefit broader applicability. utilization of preference-aware embeddings significantly enhances recommendation systems but does not fully address the dynamic state modeling required for better generalization across more extensive graph-based applications \cite{li2023preference}. These findings highlight the need for a model that dynamically adapts to changing node states while taking the edge features into account, enhancing performance and utility for a range of graph-based applications. 
        Our work combines spectral and non-spectral approaches to graph convolution of \cite{hamilton2020inductive} and \cite{kipf2017semi}. Specifically, the utilization of spectral methods is similar to the formulations by \cite{shuman2020emerging} and \cite{lee2021graph}, where the adjacency and laplacian matrices are used to capture graph properties in the structural aspect. The spectral techniques are extended by incorporating them with the non-spectral methods that process features adaptively based on the local context without the need to know the global structure of the graph. This enhances the ability to discern and adapt to complex patterns in graph data and improves scalability and flexibility in handling diverse graph sizes and topologies. S3M dynamically operates on specific states of the system to enhance prediction accuracy. On the other hand, attention mechanisms help to maintain contextual awareness, which aids in improving the model's relevance and precision. Inspired by the advancement, a selective state-space-based attention architecture is proposed to operate on graph data. The idea is to emphasize the important links of the graph, taking the node states into account. This architecture has several interesting properties: (1) enhanced feature extraction and representation, with each layer focusing on key node states and connections while adapting to changing node states; (2) scalable by managing large, dynamic graphs while preserving node characteristics; (3) better generalization to unseen graph structures by focusing on relevant node interactions and adapting to new structures. The proposed architecture is validated on four challenging benchmarks: \textit{Cora} \cite{cora}, \textit{Citeseer} \cite{citeseer} and \textit{Pubmed} \cite{pubmed} citation networks and \textit{protein-protein interaction} (PPI) dataset \cite{hamilton2017inductive}, outperforming major state-of-the-art results displaying the potential of Graph Selective States Focused Attention Networks (GSANs) in handling graph-structured data.
\subsection{Related Work}

GNNs are foundational architectures when addressing the complexities inherent in graph-structured data across various domains, from transportation systems to supply chain networks. Traditionally, transductive learning methods, including GCNs \cite{kipf2017semi} and GAT \cite{velickovic2018graph}, have utilized the graph structure for propagating node features and labels effectively. These methods, including others like Node2Vec \cite{grover2016node2vec} and DeepWalk \cite{perozzi2014deepwalk}, use adjacency matrices and local neighborhood features to learn node embeddings. While effective, they often struggle with scalability and adaptability to dynamic graph changes.

On the inductive learning front, techniques like GraphSAGE \cite{hamilton2017inductive} and graph isomorphism networks (GIN) \cite{xu2018powerful} aim to generalize beyond the training data to unseen nodes or entirely new graphs. These methods employ strategies like feature aggregation from fixed-size neighborhoods, allowing them to handle dynamic graph structures effectively. For instance, \cite{cao2023improving} introduced GSTAN for dynamic point-of-interest recommendations by integrating trajectory learning and enhancing spatial-temporal data analysis. Similarly, \cite{li2024stgmamba} and \cite{xu2023intelligent} have integrated state space models and deep reinforcement learning into GNNs to better capture temporal dynamics and improve decision-making in dynamic environments.

Despite this progress, challenges remain in dynamically adjusting to node states and efficiently handling large-scale graphs. Our work introduces GSAN, which uses both spectral and non-spectral GNN approaches but also integrates S3M and graph attention mechanisms. This novel combination enhances node interaction representation, improves model adaptability to graph dynamics, and boosts performance across various graph-based tasks. Furthermore, the integration of these into GSAN provides an efficient framework for dealing with the non-Euclidean nature of graphs, which traditional methods struggle with. By focusing on both node and edge features, GSAN enables a better understanding and processing of graph data, which ensures that both the attributes of individual nodes and their connections are utilized optimally for classification and prediction tasks, outperforming existing models in both accuracy and generalization on standard benchmarks like \textit{Cora}, \textit{Citeseer}, and \textit{PubMed}. Our model's ability to adjust to changing node states dynamically and generalize to unseen graph structures offers a solution for a vast range of applications which include traffic management, social network analysis, drug discovery, knowledge graphs, electric power grids, etc.

\section{GSAN Architecture}

\begin{figure*}
    \centering
    \includegraphics[width=1\linewidth, height=0.1655\linewidth]{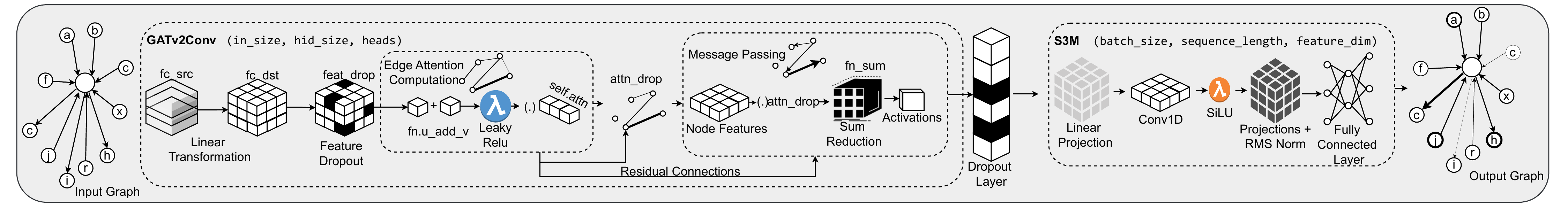}
    \caption{GSAN Architecture Diagram}
    \label{fig:arch}
\end{figure*}

This section introduces the building block layer used to create S3M networks through layer stacking. The Architecture is illustrated in Fig. \ref{fig:arch}. GSAN operates graphs with node features as input, initially passed through the \textit{GATv2Conv} layer \cite{brody2022how}. We utilize this layer to implement a graph attention mechanism where edge features are first linearly transformed and then used to compute attention coefficients that determine the significance of neighbor nodes' links. These coefficients are obtained by applying a \textit{LeakyReLU} activation on the sum of transformed features of neighboring node pairs, followed by a \textit{softmax} to ensure they sum to one, signifying their relative importance. After attention scores are applied to the corresponding features, they are aggregated to form the updated node representations. These representations are then passed through a dropout layer for regularization and subsequently fed into the \textit{S3Block}. This block applies additional transformations, including a convolution and a custom selective scan, integrating and enhancing node feature interactions further. The output from this block is normalized and finally projected through a linear layer to produce the final output, which is suitable for downstream tasks like classification or regression.
:
\begin{algorithm}[!ht]
\caption{GSAN Training Procedure}
\label{algorithm:GSAN}
\textbf{Input:} Node features $\mathbf{X} \in \mathbb{R}^{N \times F}$, adjacency matrix $\mathbf{A} \in \mathbb{R}^{N \times N}$ \\
\textbf{Output:} A transformed feature matrix $\mathbf{Y}_T$ for graph analysis, combining attention and state space models.
\begin{algorithmic}[1]
    \For{$h = 1$ \textbf{to} $\text{heads}$}
    
    \State \hspace{-0.3cm} $\mathbf{Z}^{(h)} = \sigma \left( \sum_{j \in \mathcal{N}(i)} \alpha_{ij}^{(h)} (W^{(h)} x_j) \right)$
    \begin{equation}
    \label{eq:Z_update}
    \end{equation}
    \State \hspace{-0.3cm} $\alpha_{ij}^{(h)} = \frac{\exp \left( \text{LeakyReLU}(a^{(h)T} [W^{(h)} x_i \| W^{(h)} x_j]) \right)}{\sum_{k \in \mathcal{N}(i)} \exp \left( \text{LeakyReLU}(a^{(h)T} [W^{(h)} x_i \| W^{(h)} x_k]) \right)}$ 
    \begin{equation}
    \label{eq:alpha_update}
    \end{equation}
\EndFor

    \State $\mathbf{Z} = \text{Concatenate}(\mathbf{Z}^{(1)}, \dots, \mathbf{Z}^{(\text{heads})})$
    \State $\mathbf{Z}_{\text{proj}} = W^{(proj)} \mathbf{Z}$
    \State Split $\mathbf{Z}_{\text{proj}}$ into $[\mathbf{Z}_1, \mathbf{Z}_2]$ along features
    \State $\mathbf{Z}_1 = \text{Conv1D}(\mathbf{Z}_1, W^{(conv)}, b^{(conv)})$
    \State $\mathbf{Z}_{\text{gate}} = \sigma(\mathbf{Z}_2)$
    \State $\mathbf{Z}_{\text{ssm}} = \mathbf{Z}_1 \odot \mathbf{Z}_{\text{gate}}$
    \State $\mathbf{Z}_{\text{out}} = W^{(out)} \mathbf{Z}_{\text{ssm}}$
    \State $\mathbf{U} \gets \mathbf{Z}_{\text{out}}$
    \For{$t = 1$ \textbf{to} $T$}
    \State \begin{equation} 
    \hspace{-3.5cm} \mathbf{X}_t = e^{-\Delta \mathbf{A}} \mathbf{X}_{t-1} + \mathbf{B} \odot \mathbf{U}
    \label{eq:X_update}
    \end{equation}
    \State \begin{equation}
    \hspace{-4.4cm} \mathbf{Y}_t = \mathbf{C} \mathbf{X}_t + \mathbf{D} \odot \mathbf{U}
    \label{eq:Y_update}
    \end{equation}
\EndFor

    \State \Return $\mathbf{Y}_T$
\end{algorithmic}
\end{algorithm}

Implementation. As shown in Algorithm \ref{algorithm:GSAN}, it begins with a \textit{GATv2} convolution that applies multiple attention heads to compute refined node interactions, allowing the model to focus selectively on influential link features while suppressing less relevant information, improving model interpretability and performance. The attention mechanism uses $\sigma$: the nonlinear activation function \textit{LeakyReLU}; $\alpha_{ij}^{(h)}$: attention coefficients between node $i$ and node $j$ for head $h$; $W^{(h)}$: the weight matrix for the $h$-th attention head; $x_i, x_j, x_k$: feature vectors of node $i$, node $j$, and node $k$, respectively; $a^{(h)}$: the learnable parameter vector for the $h$-th attention head; and Concatenate: the operation to combine tensors along a specified dimension. The attention coefficients $\alpha_{ij}^{(h)}$ are computed as shown in Equation \ref{eq:alpha_update} and normalized using \textit{LeakyReLU} and \textit{softmax} functions, which help stabilize learning by managing gradient flow. Subsequent feature projection and splitting prepare the data for dual pathway processing, where one stream is convolved for feature transformation using $W^{(proj)}$: the projection weight matrix, and $W^{(conv)}$, $b^{(conv)}$: the weight matrix and bias vector for convolution operation. The other stream serves as a gating mechanism, effectively regulating the integration of new information and maintaining the balance between adaptability and stability in feature representation. In the state-space modeling part of the network, $\mathbf{U}$ is the updated node representation matrix for selective state modeling, while $\Delta$, $\mathbf{B}$, $\mathbf{C}$, and $\mathbf{D}$ are trainable parameters for the state-space model. $\mathbf{X}_0$ is the initial state for the state-space model, typically set to zero. The state space memory (SSM) layer updates node states iteratively, as shown in Eqs. \ref{eq:X_update} \& \ref{eq:Y_update}, capturing dynamic changes over graph structures. This allows the handling of complex dependencies within the data, making it effective for node classification and graph segmentation. The overall computational complexity of the network is driven by several key components. The \textit{GATv2} convolution is a significant contributor, with a complexity of \(O(E \cdot F)\), where \(E\) represents the number of edges in the graph and \(F\) is the feature dimension. This complexity is derived from the attention mechanism that requires computation across each edge for the feature updates. Normalization steps, which include \textit{LeakyReLU} and \textit{softmax}, have a complexity of \(O(N)\), where \(N\) is the number of nodes. These operations are generally efficient and contribute minimally to the overall complexity. The dual pathway processing involves both feature transformation and gating mechanism. Feature transformation through convolution operations typically has a complexity of \(O(N \cdot F^2)\), assuming a fully connected layer, where \(N\) is the number of nodes and \(F\) is the feature dimension. The gating mechanism, which integrates new information, shares a similar complexity, adding to the computational load. The SSM update step is iterative and has a complexity of \(O(B \cdot L \cdot D \cdot K)\), where \(B\) is the batch size, \(L\) is the sequence length, \(D\) is the input dimension, and \(K\) is the number of states. This step is crucial for capturing dynamic changes and dependencies within the graph structure. Overall, the total complexity of the network can be approximated by summing the complexities of the major components:
\[
O(E \cdot F) + O(N) + O(N \cdot F^2) + O(B \cdot L \cdot D \cdot K)
\] which simplifies to:
\[
O(E \cdot F + N \cdot F^2 + B \cdot L \cdot D \cdot K).
\]
This approximation combines the major terms where \(E \cdot F\) accounts for the \textit{GATv2} convolution complexity, \(N \cdot F^2\) covers the feature transformation and gating mechanism, and \(B \cdot L \cdot D \cdot K\) represents the iterative updates in the SSM.

\subsection{Graph Attentional Layer}
The Graph Attentional Layer (GAL) within our network uses the \textit{GATv2}, designed to enhance link feature representation through a self-attention mechanism. Node features are aggregated using attention scores to refine each node's feature set by focusing on important links. The update rule is presented in Eq. \ref{eq:Z_update}, \(\mathbf{Z}^{(h)}\) denotes the feature vector of node \(i\) at layer \((h+1)\), where \(W^{(h)}\) is the trainable weight matrix specific to the layer and \(\sigma\) represents the activation function. In Eq. \ref{eq:alpha_update}, the attention coefficients \(\alpha_{ij}^{(h)}\) for nodes \(i\) and \(j\) in layer \(h\) are determined by applying a softmax function to the exponentiated values derived from the \textit{LeakyReLU} activation function. These values result from the dot product between a learnable projection vector \(\mathbf{a}^{(h)}\) and the concatenated features of nodes \(i\) and \(j\), transformed by the weight matrix \(W^{(h)}\), where \(\|\) denotes concatenation. This mechanism ensures that the coefficients are normalized across all neighboring nodes \(k \in \mathcal{N}(i)\), as specified in Eq. \ref{eq:alpha_update}, enabling the layer to effectively process and adapt to the complexities inherent in diverse graph-structured data.

\subsection{Selective State Space Modeling Layer}

The Selective State Space Modeling (S3M) layer introduces node state information processing. This layer models the changing of node states over time, enabling the network to capture and utilize temporal dynamics effectively. The state-space representation of a linear system is captured by two fundamental equations Eq. \(\ref{eq:X_update}\) describes the system's state evolution,  Eq. \(\ref{eq:Y_update}\) shows how the output \(\mathbf{Y}_t\) is determined by the current state \(\mathbf{X}_t\) and the input \(\mathbf{U}\), influenced by the output matrix \(\mathbf{C}\) and the direct transmission matrix \(\mathbf{D}\) in Eq. \ref{eq:Y_update} Here, the output \(\mathbf{Y}_t\) depends on the current state \(\mathbf{X}_t\) and the input \(\mathbf{U}\), modified by the output matrix \(\mathbf{C}\) and the direct transmission matrix \(\mathbf{D}\). Eqs. \ref{eq:X_update} and \ref{eq:Y_update}, together, define the dynamics and output of the system, showing how inputs and states translate into system outputs over time. In the context of our network, the S3M layer processes node features as state vectors where \(\mathbf{X}_t\) represents the state vector of a node at time \(t\), and \(\mathbf{U}\) encapsulates relevant input features or external signals affecting the node at the same time. The output \(\mathbf{Y}_t\) denotes the transformed state, which could represent the node's predictions, classifications, or enhanced features for subsequent layers. The matrices \(\mathbf{A}\), \(\mathbf{B}\), \(\mathbf{C}\), and \(\mathbf{D}\) are adaptive parameters learned during training, defining the transition and output dynamics of the system. The S3M layer selectively adjusts these parameters based on input characteristics and node interactions within the graph, employing advanced mechanisms such as parameter sharing, which enhances learning efficiency and model accuracy. This capability enables the layer to manage node state effectively transitions over time, making it particularly suited for tasks such as dynamic node classification, temporal link prediction, or anomaly detection in networks where understanding the evolution of node states is crucial.

\section{Experimental Study}

This section summarizes our experimental setup, datasets used, and a comparative performance analysis of GSAN model. The comparative evaluation of GSAN models is performed against a variety of strong baselines and previous approaches on four established graph-based benchmark tasks (transductive as well as inductive), matched or outperformed state-of-the-art performance across most of them. We utilized a configuration of 48 \textit{vCPUs}, 64 GB RAM on \textit{Ubuntu}, and Nvidia's Tesla V100S-PCIE-32GB \textit{GPU} to generate our results. For both the transductive and inductive setups, hyperparameters included a learning rate of 0.005, weight decay of 5e-4, 8 attention heads (for \textit{GAT}), 8 hidden units, and a dropout rate of 0.6. Models were trained using the \textit{Adam} optimizer with regularization techniques such as \textit{L1} regularization, \textit{L2} regularization, and \textit{Smooth L1 loss}, using early stopping with a patience factor of 10 epochs.

\textbf{Transductive Learning Setup:} We evaluated state-of-the-art models, including GCNs \cite{kipf2017semi}, GAT \cite{velickovic2018graph}, AMC-GCN \cite{wang2020amc}, NIGCN \cite{wang2021nigcn}, DropEdge \cite{Rong2020DropEdge}, NodeNorm \cite{zhao2020node}, GDC \cite{klicpera2019diffusion}, CensNet \cite{he2020censnet}, NENN \cite{chen2020neural}, EGAT \cite{zhang2020egat}, etc. and our GSAN model on citation network datasets like \textit{Cora}, \textit{Citeseer}, and \textit{Pubmed}, using classification accuracy as the metric, reflecting effectiveness through accurate node classification based on learned representations.

\textbf{Inductive Learning Setup:} We assessed models such as GraphSAGE \cite{hamilton2017inductive}, GIN \cite{xu2018powerful}, FastGCN \cite{chen2018fastgcn}, Cluster-GCN \cite{chiang2019cluster}, PinSage \cite{ying2018graph}, etc. and our GSAN model on datasets like \textit{PPI} networks, measuring performance with accuracy and \textit{F1 score}, as the model must generalize to unseen nodes or entirely new graphs.

\subsection{Datasets}
\textbf{Transductive Learning Datasets}: The three citation network benchmark dynamic graph library datasets are used: \textit{Cora}, \textit{Citeseer}, and \textit{Pubmed}. In all these datasets, nodes correspond to documents and edges to (undirected) citations. Node features correspond to elements of a bag-of-words representation of a document. Each node has a class label. The efficiency of the model is evaluated on 1000 test nodes, and 500 additional nodes are used for validation purposes. The \textit{Cora} dataset contains 2708 nodes, 10556 edges, 7 classes, and 1433 features per node. The \textit{Citeseer} dataset contains 3327 nodes, 9288 edges, 6 classes, and 3703 features per node. The \textit{Pubmed} dataset contains 19717 nodes, 88615 edges, 3 classes, and 500 features per node.

\textbf{Inductive Learning Datasets}: Protein-protein interaction (PPI) dataset is used that consists of graphs corresponding to different human tissues. The dataset contains 20 graphs for training, 2 for validation, and 2 for testing. The testing graphs are unobserved during training. The average number of nodes per graph is 2371. Each node has 50 features. that is composed of positional gene sets, motif gene sets, and immunological signatures; there are 121 labels for each node set from gene ontology collected from the molecular signatures database, and a node can possess several labels simultaneously.

\begin{table}[!ht]
\setlength{\tabcolsep}{1mm}
    \centering
    
    \begin{tabular}{lccc}
        \hline
        \textbf{Method} & \textbf{Cora} & \textbf{Citeseer} & \textbf{Pubmed} \\
        \hline
        MLP & 55.1\% & 46.5\% & 71.4\% \\
        ICA \shortcite{Lu2003LinkBasedClassification} & 75.1\% & 69.1\% & 73.9\% \\
        LP \shortcite{Zhu2003SemiSupervisedLG} & 68.0\% & 45.3\% & 63.0\% \\
        ManiReg \shortcite{Belkin2006ManifoldR} & 59.5$\pm$0.2 &  60.1$\pm$0.3 & 70.7$\pm$0.4 \\
        SemiEmb \shortcite{Weston2012DeepLearning} & 59.0\% & 59.6\% & 71.7\% \\
        DeepWalk \shortcite{perozzi2014deepwalk} & 67.2\% & 43.2\% & 65.3\% \\
        Monet \shortcite{Monti2016GeometricDL} & 81.7$\pm$0.5\% & — & 78.8$\pm$0.3\% \\
        Planetoid \shortcite{Yang2016SemiSupervisedGE}  &  75.7 & 64.7 & 77.2 \\
        Chebyshev \shortcite{Defferrard2016CNNOnGraphs} &  81.2 & 69.8 & 74.4 \\
        GCN \shortcite{Kipf2017GCN} & 81.5$\pm$0.5 & 70.4$\pm$0.4 & 78.1$\pm$0.4 \\
        GAT \shortcite{velickovic2018graph} & 83.0$\pm$0.5 & 71.6$\pm$0.8 & 78.2$\pm$0.4 \\
        JKNet \shortcite{xu2018representationlearninggraphsjumping} & 76.4$\pm$2.5\% & 62.6$\pm$3.4\% & 77.3$\pm$0.6\% \\
        GDC \shortcite{Klicpera2019DiffusionGNN} & \underline{83.8$\pm$0.2} & 73.3$\pm$0.3 & 79.9$\pm$0.3 \\
        CensNet \shortcite{Jiang2019CensNet} & 79.4$\pm$1.0 & 62.5$\pm$1.5 & 69.9$\pm$2.1 \\
        SGC \shortcite{pmlr-v97-wu19e} & 78.3$\pm$1.0\% & 69.0$\pm$0.9\% & 75.4$\pm$1.7\% \\
        DropEdge (64) \shortcite{Rong2020DropEdge} & 78.9$\pm$0.3 & 65.1$\pm$0.5 & 76.9$\pm$0.6 \\
        NENN \shortcite{Yang2020NENN} & 82.6$\pm$0.1 & 68.2$\pm$0.1 & 77.7$\pm$0.1 \\
        SIGN \shortcite{frasca2020signscalableinceptiongraph} & 81.7$\pm$1.0\% & 69.1$\pm$0.8\% & 79.6$\pm$1.5\% \\
        ADSF \shortcite{Zhang2020AdaptiveSF} & 82.3$\pm$0.8\% & 69.1$\pm$1.8\% & 80.0$\pm$0.7\% \\
        EGAT \shortcite{Wang2021EGAT} & 82.1$\pm$0.7 & 70.3$\pm$0.5 & 78.1$\pm$0.4\\
        $S^{2}$GC \shortcite{Zhu2021SimpleSG} & 82.1$\pm$0.4\% & 68.9$\pm$0.9\% & 79.8$\pm$0.6\% \\
        SPAGAN \shortcite{yang2021spaganshortestpathgraph} & 82.0$\pm$0.7\% & 69.0$\pm$1.5\% & 79.5$\pm$0.5\% \\
        NodeNorm (64) \shortcite{Zhou2021PerformanceDegradationGNN} & 83.4$\pm$0.6 & \underline{73.8$\pm$0.8} & 80.4$\pm$1.2 \\
        AMC-GCN \shortcite{Shi2022AdaptiveGNN} & 83.4$\pm$0.4 & 72.8$\pm$0.5 & 78.9$\pm$0.3 \\
        APPNP \shortcite{gasteiger2022predictpropagategraphneural} & 82.1$\pm$1.4\% & 69.2$\pm$1.3\% & 79.6$\pm$1.7\% \\
        NIGCN \shortcite{Huang2023NodeWiseDiffusion} & 82.1$\pm$1.1 & 71.4$\pm$0.8 & 80.9$\pm$2.0 \\
        HONGAT \shortcite{Zhang2024HONGATGA} & 83.1$\pm$1.0\% & 69.5$\pm$1.2\% & \underline{81.1$\pm$0.8\%} \\
        \textbf{GSAN (Ours)} & \textbf{84.7$\pm$0.3} & \textbf{80.4$\pm$1.0} & \textbf{81.4$\pm$1.2} \\
        \hline
    \end{tabular}
    \caption{Comparison of node classification accuracy on inductive task with other GNN methods (highest accuracy highlighted in bold and second highest is underlined).}
    \label{tab:comparison}
\end{table}

\begin{table}[!ht]
\setlength{\tabcolsep}{1mm}
\centering
\begin{tabular}{@{}lc@{}}
\hline
\textbf{Method}                          & \textbf{F1 Score}      \\ 
\hline
Random                                   & 0.396                  \\
MLP                                      & 0.422                  \\
GraphSAGE-GCN \shortcite{Hamilton2017InductiveRL}    & 0.500                  \\
GraphSAGE-mean \shortcite{Hamilton2017InductiveRL}   & 0.598                  \\
GraphSAGE-LSTM \shortcite{Hamilton2017InductiveRL}    & 0.612                  \\
GraphSAGE-pool \shortcite{Hamilton2017InductiveRL}   & 0.600                  \\
GraphSAGE (Best)                               & 0.768                  \\
Const-GAT \shortcite{velickovic2018graph}                        & \(0.934 \pm 0.006\)    \\
GAT \shortcite{velickovic2018graph}                               & \underline{\(0.973 \pm 0.002\)}    \\ 
\textbf{GSAN (Ours)} & \(\textbf{0.988} \pm \textbf{0.001}\) \\

\hline
\end{tabular}
\caption{Inductive method performance on PPI dataset}
\label{tab:ppi_performance}
\end{table}

\begin{figure*}[!ht]
    \centering
    \begin{subfigure}{0.32\textwidth}
        \centering
        \includegraphics[width=\linewidth, height=0.65\textwidth]{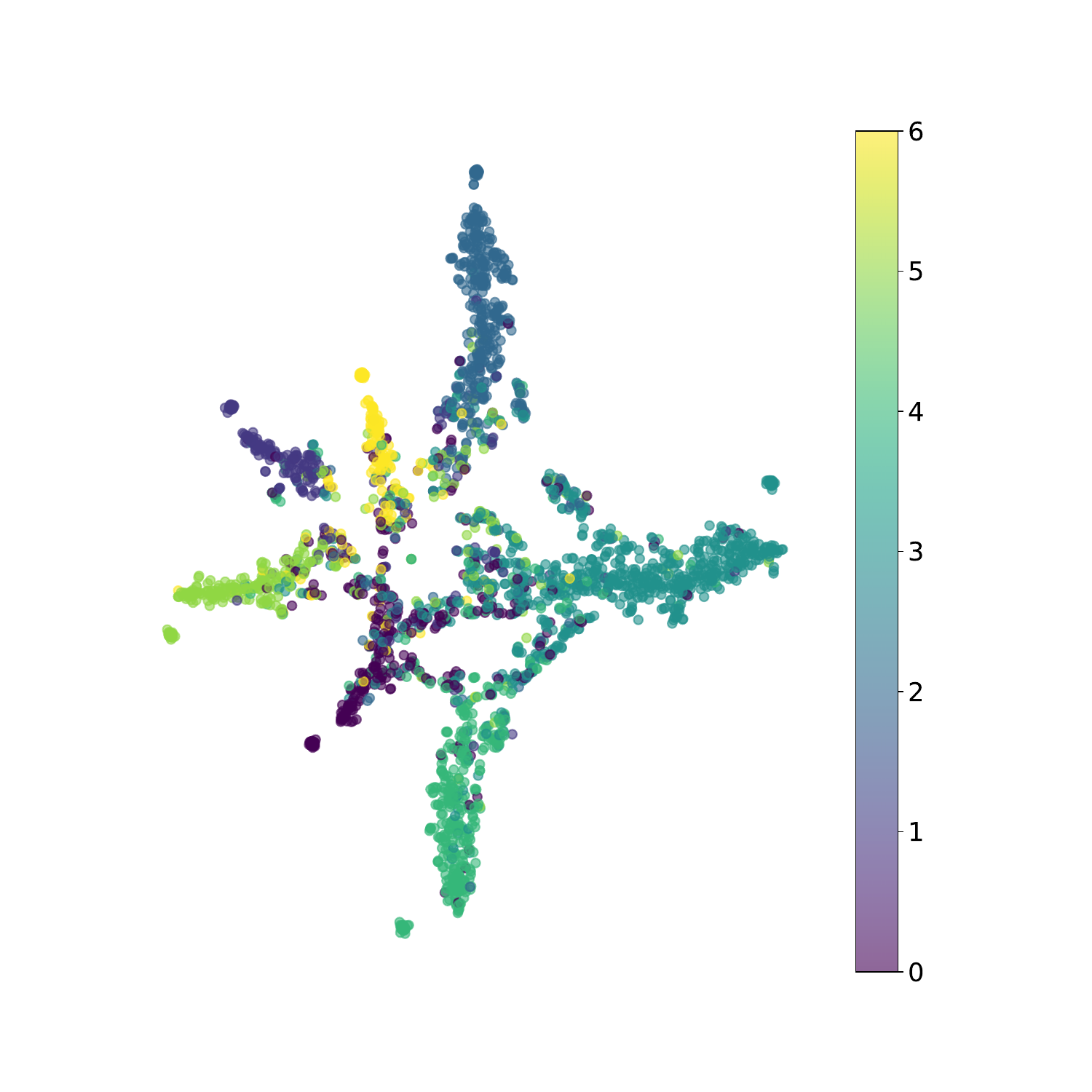}
        \caption{Cora}
        \label{fig:cora_tsne}
    \end{subfigure}\hfill
    \begin{subfigure}{0.32\textwidth}
        \centering
        \includegraphics[width=\linewidth, height=0.65\textwidth]{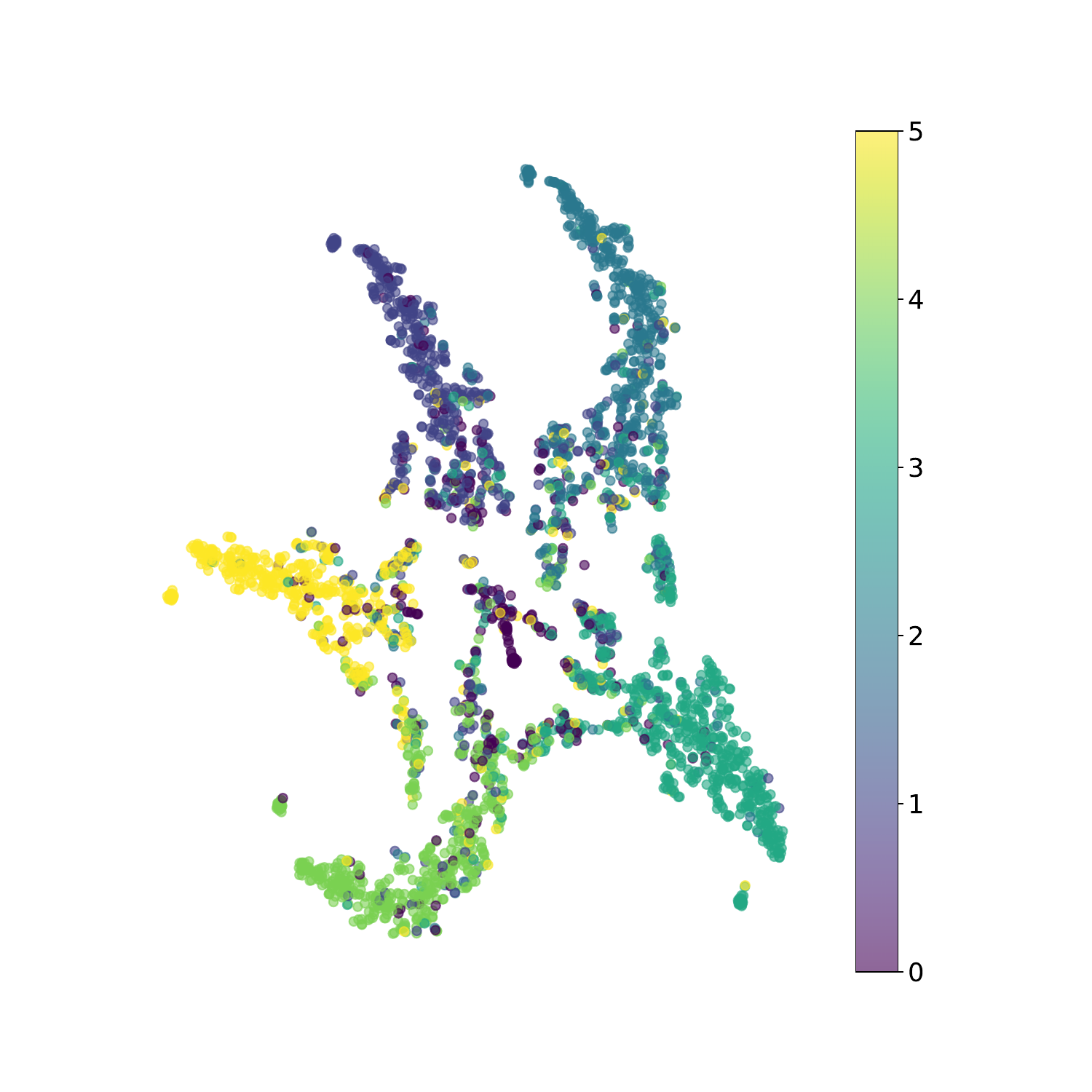}
        \caption{Citeseer}
        \label{fig:citeseer_tsne}
    \end{subfigure}\hfill
    \begin{subfigure}{0.32\textwidth}
        \centering
        \includegraphics[width=\linewidth, height=0.65\textwidth]{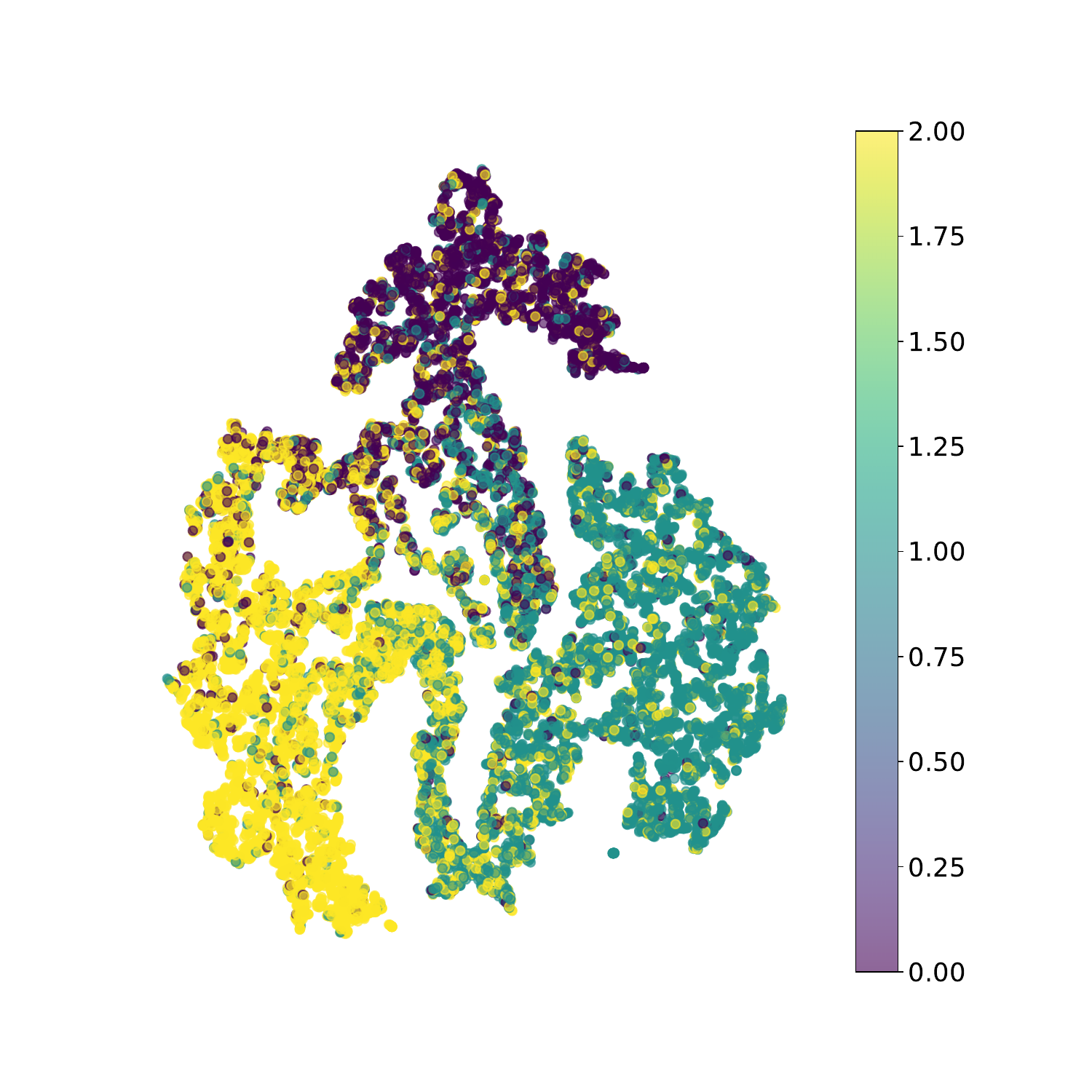}
        \caption{Pubmed}
        \label{fig:pubmed_tsne}
    \end{subfigure}
    \caption{TSNE (t-Distributed Stochastic Neighbor Embedding) Plot for the datasets}
    \label{fig:tsne}
\end{figure*}

\begin{figure*}[!ht]
    \centering
    \begin{subfigure}{0.32\textwidth}
        \centering
        \includegraphics[width=1.15\linewidth, height=0.7\linewidth]{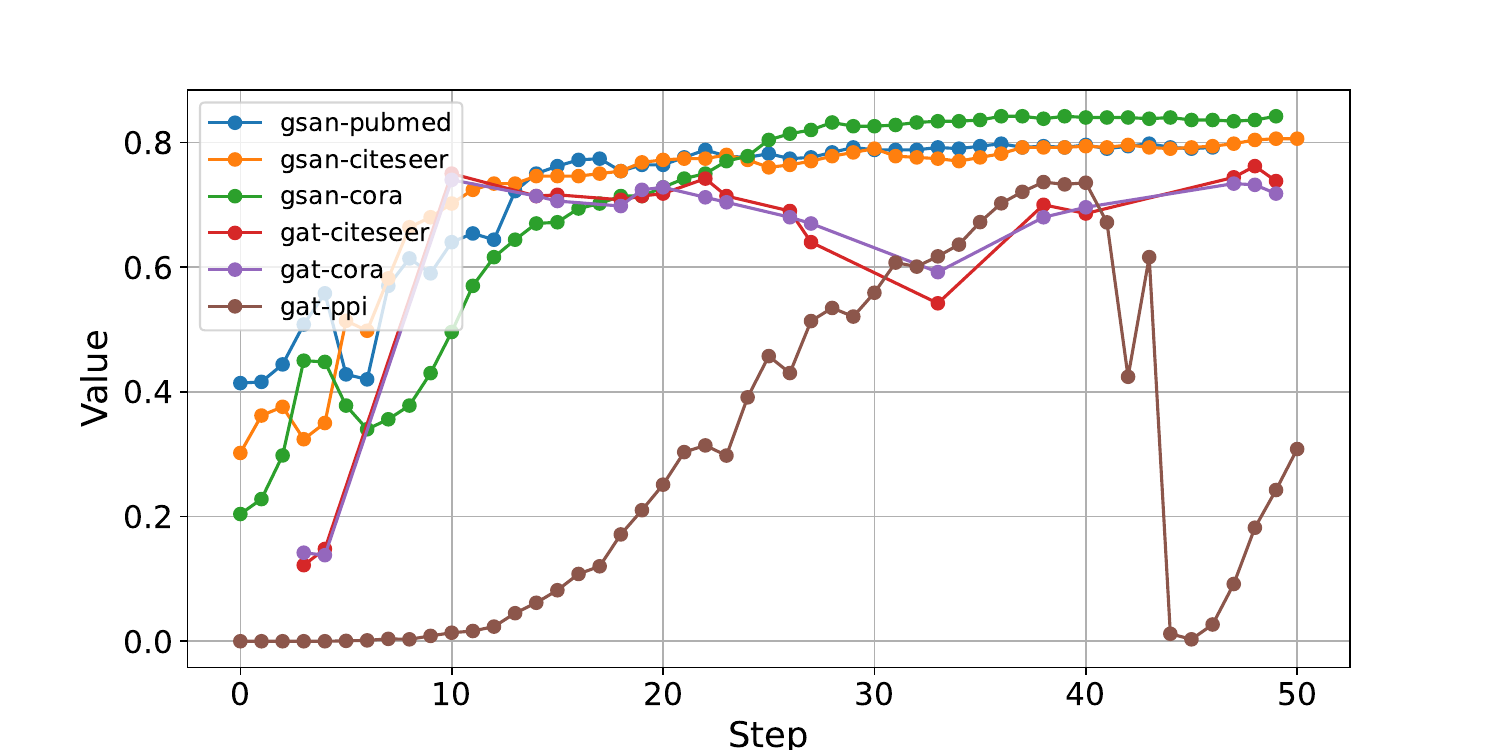}
        \caption{Validation accuracies}
        \label{fig:val_acc}
    \end{subfigure}\hfill
    \begin{subfigure}{0.32\textwidth}
        \centering
        \includegraphics[width=1.15\linewidth, height=0.7\linewidth]{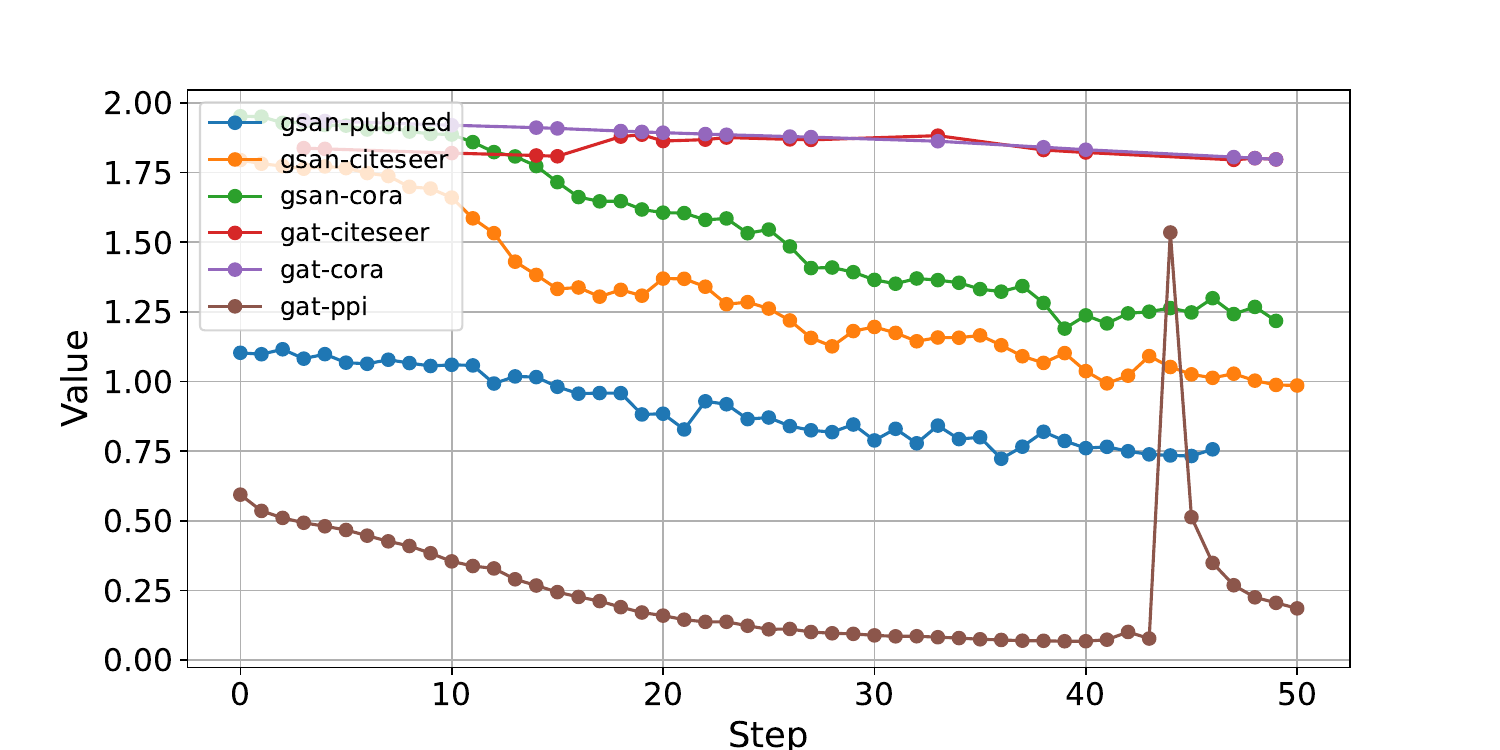}
        \caption{Validation loss}
        \label{fig:val_loss}
    \end{subfigure}\hfill
    \begin{subfigure}{0.32\textwidth}
        \centering
        \includegraphics[width=1.15\linewidth, height=0.7\linewidth]{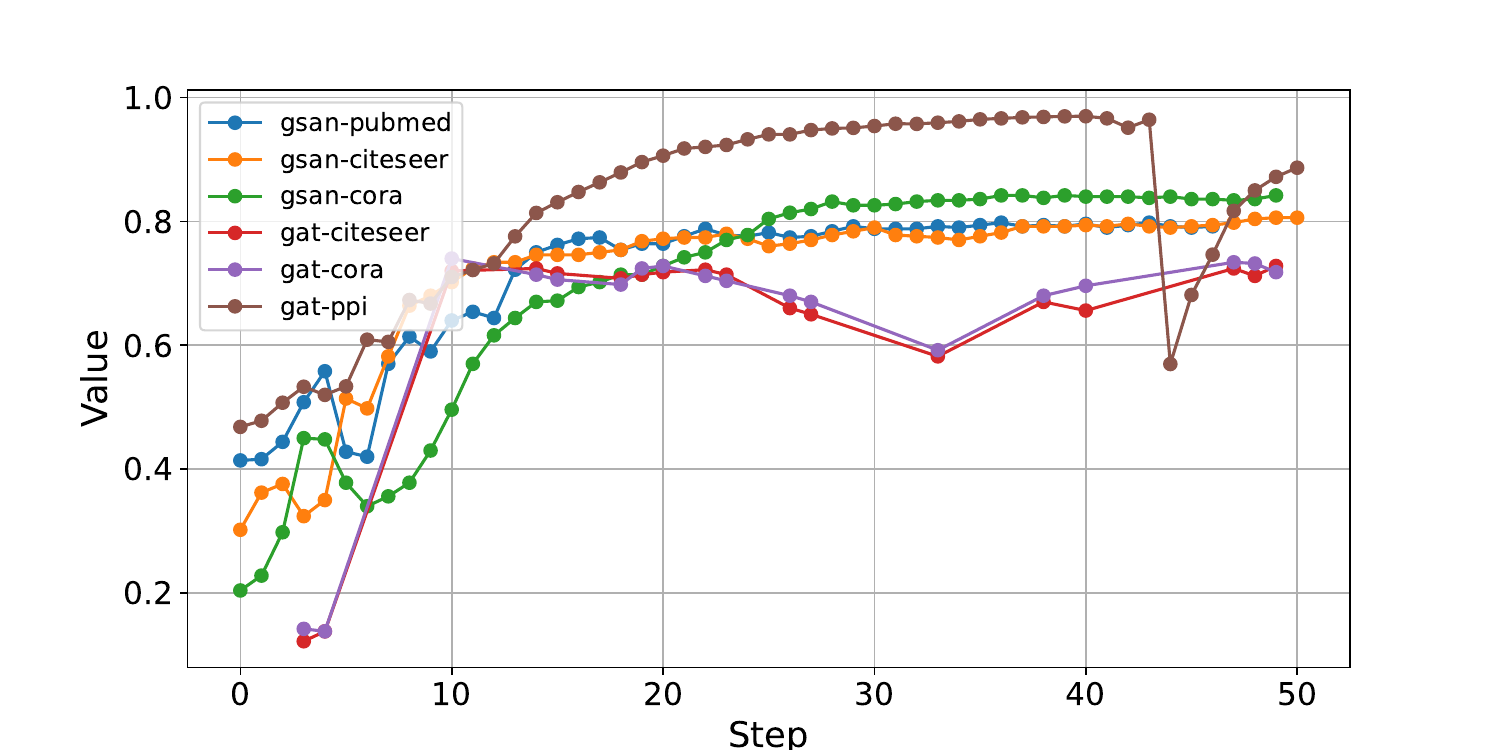}
        \caption{Validation F1 score}
        \label{fig:val_micro_f1}
    \end{subfigure}
    \caption{Comparison of validation metrics across different classification tasks.}
\end{figure*}

\begin{figure*}[!ht]
    \centering
    \begin{subfigure}{0.33\textwidth}
        \centering
        \includegraphics[width=1.15\linewidth, height=0.7\linewidth]{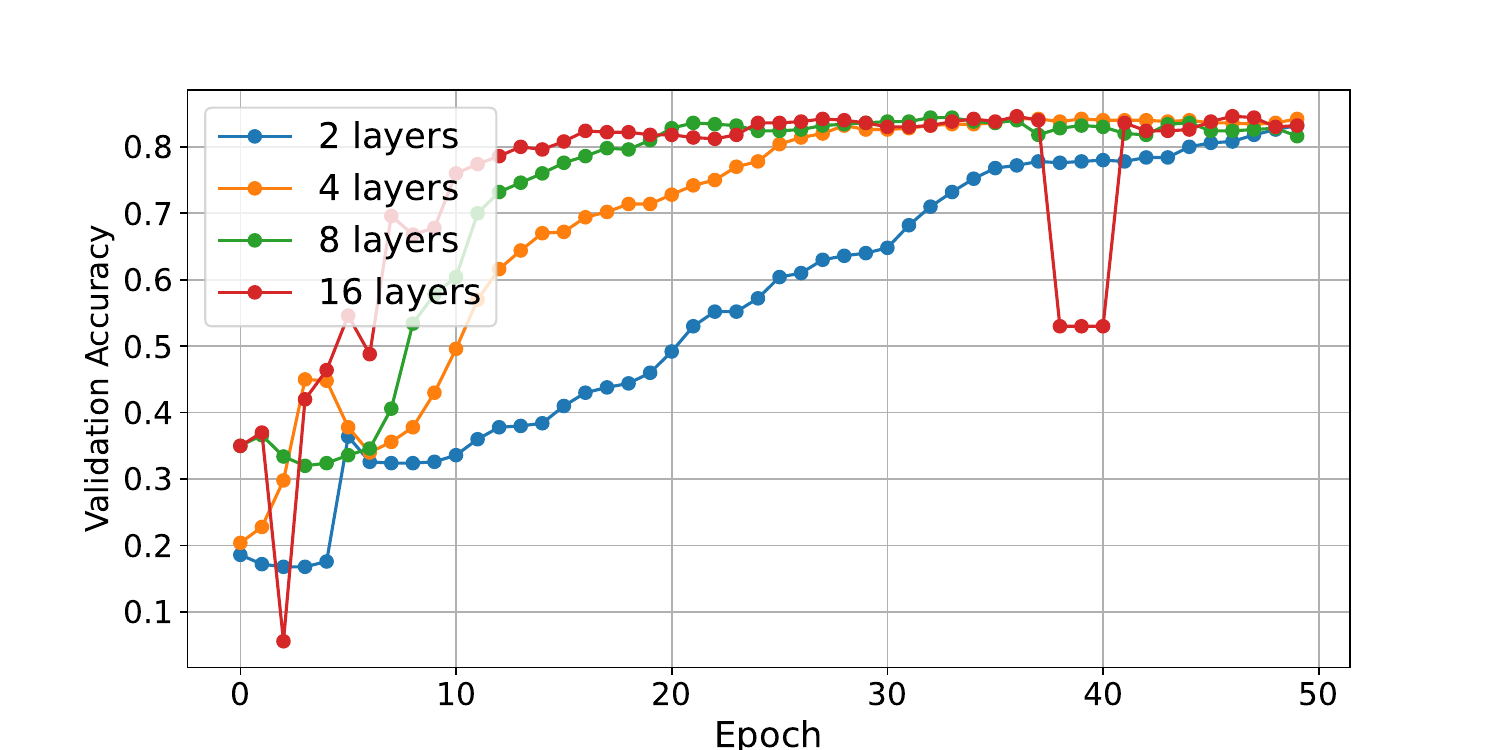}
        \caption{Validation Accuracy for Cora}
        \label{fig:acc_cora}
    \end{subfigure}\hfill
    \begin{subfigure}{0.33\textwidth}
        \centering
        \includegraphics[width=1.15\linewidth, height=0.7\linewidth]{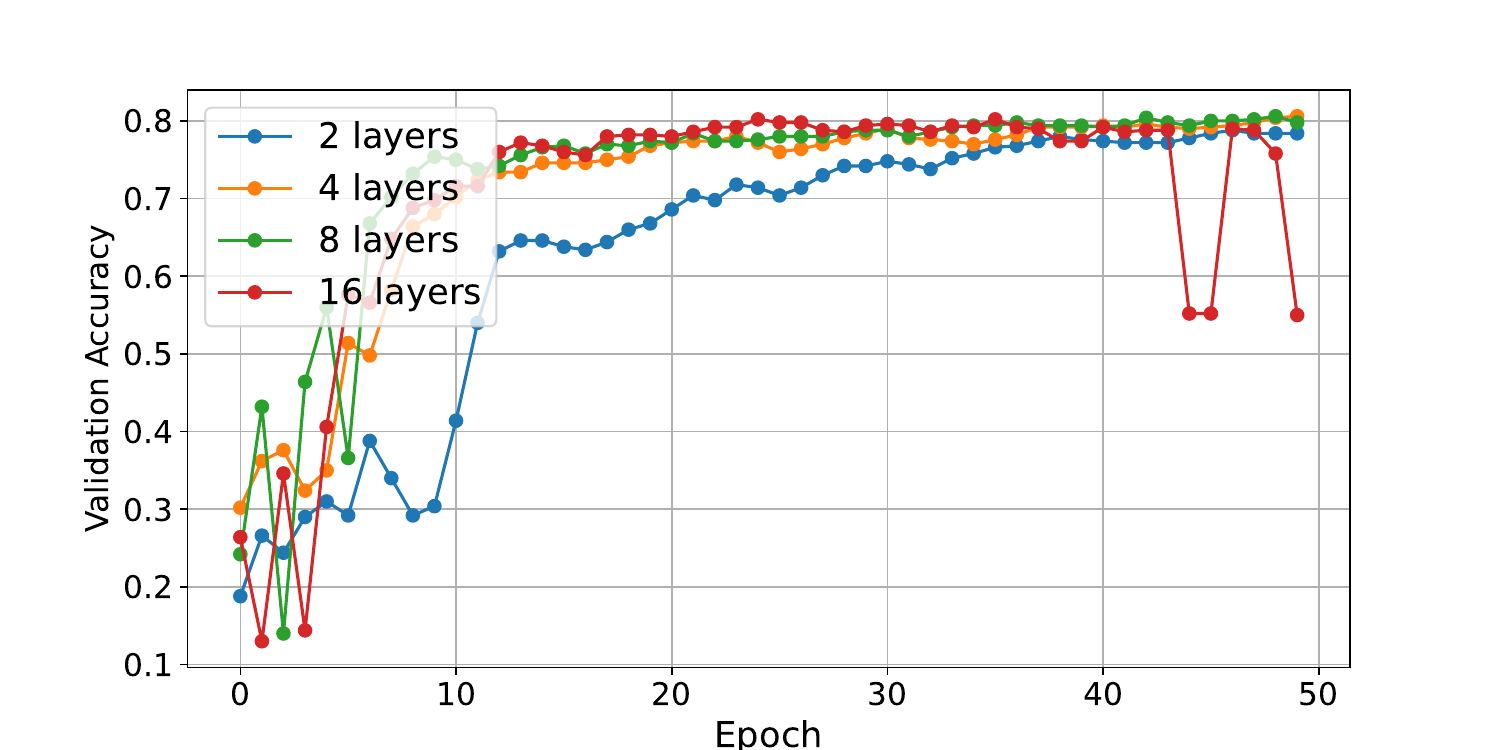}
        \caption{Validation Accuracy for Citeseer}
        \label{fig:acc_citeseer}
    \end{subfigure}\hfill
    \begin{subfigure}{0.33\textwidth}
        \centering
        \includegraphics[width=1.15\linewidth, height=0.7\linewidth]{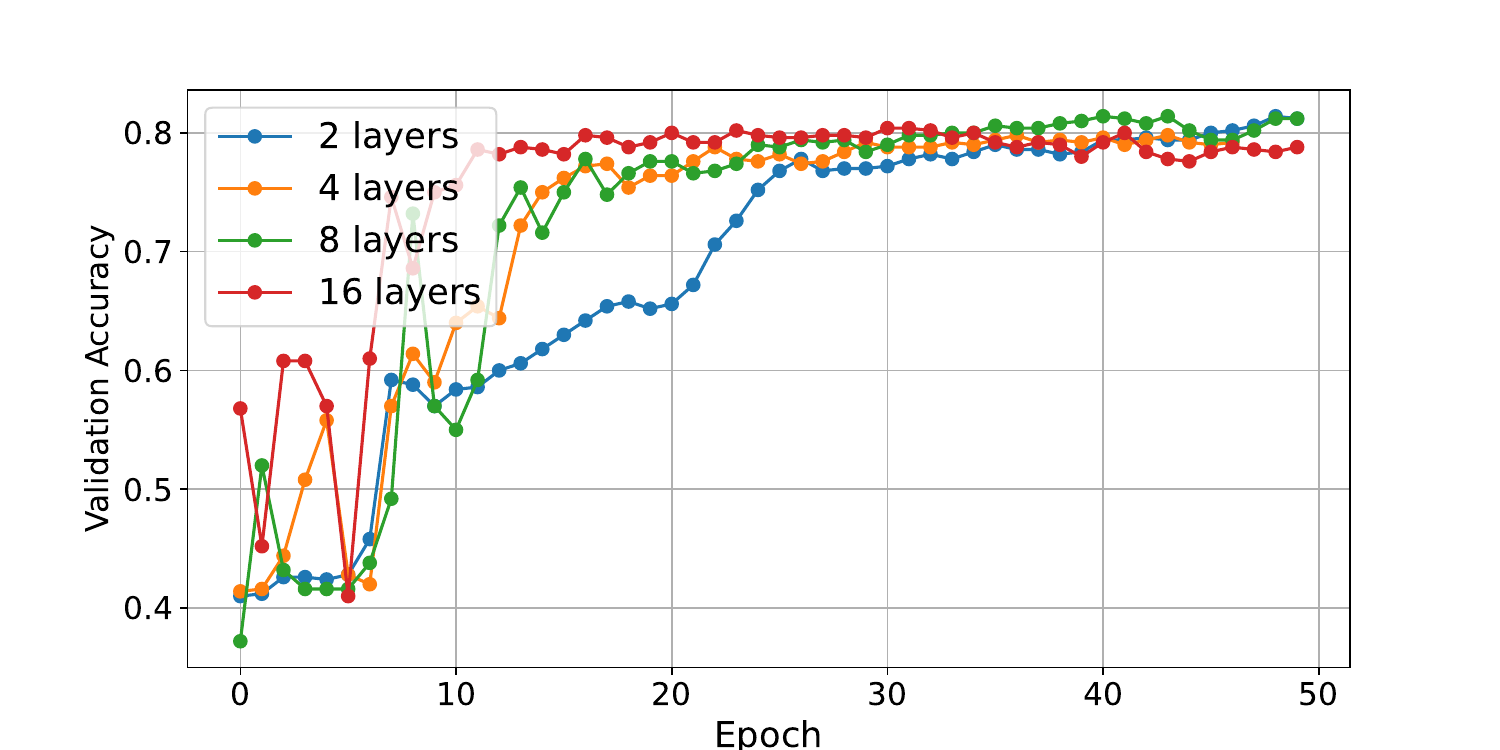}
        \caption{Validation Accuracy for Pubmed}
        \label{fig:acc_pubmed}
    \end{subfigure}
    \caption{Validation accuracy across the network for different datasets with varying layers}
\end{figure*}

\subsection{GSAN Comparative Performance Analysis}

   This section discusses the comparative performance analysis of the model. The plots for T-distributed stochastic neighbor embedding (TSNE) in Fig. \ref{fig:tsne} illustrate the ability of GSAN to classify the nodes efficiently. The validation accuracies across different models in a classification task are outlined in Fig. \ref{fig:val_acc} and validation loss in Fig. \ref{fig:val_loss} and F1 score in Fig. \ref{fig:val_micro_f1}. The validation accuracies across the various datasets with varying 2, 4, 8, and 16 layers are compared in Fig. \ref{fig:acc_cora}, Fig. \ref{fig:acc_citeseer} and Fig. \ref{fig:acc_pubmed}, on transductive tasks after 20 runs. For the inductive task, the averaged F1 score is reported on the nodes of two unseen test graphs averaged after 15 runs, and we reuse the metrics already reported in \cite{Xu_Zhang_Bian_Dwivedi_Ke_2024, Zhang_Zhang_Zhou_Li_2024, Zhou2023Nature} for the other techniques. The Table \ref{tab:comparison} \& \ref{tab:ppi_performance} present node classification accuracy and F1 score of various GNN methods on \textit{Cora}, \textit{Citeseer}, \textit{Pubmed}, and \textit{PPI} datasets respectively. It shows that GSAN (Ours) achieves the highest accuracy across all datasets. Other methods, such as GAT \cite{velickovic2018graph} and GCN \cite{Kipf2017GCN}. Specifically, GSAN achieves a max accuracy of 85\% on the \textit{Cora} dataset, 81.4\% on the \textit{Citeseer} dataset, 82.6\% on the \textit{Pubmed}, and F1 score of 0.989 on \textit{PPI} dataset marking improvements of 1.56\%, 8.94\%, 0.37\% and 1.54\% respectively over the next best performing models. The results demonstrate that our GSAN model outperforms other SOTA models across all four datasets. These results highlight the efficacy of the mechanism in improving the representation and classification capabilities of GNNs. These improvements are attributed to the architecture that enhances the representation and classification capabilities of GNNs by dynamically learning node states while effectively managing interactions in varying contexts. This allows the network to capture temporal dynamics and interactions more accurately, thus improving generalization capabilities on unseen graph structures and improving the interpretability of how node states influence link importance. This approach contributes significantly to the improved performance metrics observed in experiments.
 
\section{Conclusion}
This paper presents GSAN, a novel neural network architecture designed for graph data, which leverages selective state space modeling and graph attention mechanisms to dynamically adjust to changing node states, taking link features into account. By integrating both spectral and non-spectral approaches, GSAN effectively captures complex graph structures and enhances feature representation, leading to improved prediction accuracy and generalization in various graph-based tasks. Our work demonstrates the superiority of GSAN over existing methods via extensive experiments on benchmark datasets, including \textit{Cora}, \textit{Citeseer}, \textit{Pubmed}, and the \textit{PPI} dataset. GSAN achieves state-of-the-art performance, significantly improving node classification accuracy. 

The proposed network is not limited to node classification, and we include these results to present an initial understanding of our network's architecture and its capabilities in handling graph-based tasks. In the future, this network has the potential to solve complex problems like new drug discovery, cancer research, traffic flow optimization, cybersecurity, fraud detection, etc, as the proposed model preserves node states while computing the edge attention.

\end{document}